\documentclass[10pt,letterpaper]{article}
\usepackage[top=0.85in,left=2.75in,footskip=0.75in]{geometry}

\usepackage{float}

\usepackage{hyperref}

\usepackage{multirow}

\usepackage{algorithm}% http://ctan.org/pkg/algorithms
\usepackage{algpseudocode}% http://ctan.org/pkg/algorithmicx

\usepackage{amsmath,amssymb}

\usepackage{changepage}

\usepackage[utf8x]{inputenc}

\usepackage{textcomp,marvosym}

\usepackage{cite}

\usepackage{nameref,hyperref}

\usepackage[right]{lineno}

\usepackage{microtype}
\DisableLigatures[f]{encoding = *, family = * }

\usepackage[table]{xcolor}

\usepackage{array}

\newcolumntype{+}{!{\vrule width 2pt}}

\newlength\savedwidth

\newcommand\thickhline{\noalign{\global\savedwidth\arrayrulewidth\global\arrayrulewidth 2pt}%
\hline
\noalign{\global\arrayrulewidth\savedwidth}}

\raggedright
\usepackage[aboveskip=1pt,labelfont=bf,labelsep=period,justification=raggedright,singlelinecheck=off]{caption}

\makeatletter
\renewcommand{\@biblabel}[1]{\quad#1.}
\makeatother

\usepackage{lastpage,fancyhdr,graphicx}
\usepackage{epstopdf}
\fancyheadoffset[L]{2.25in}
\fancyfootoffset[L]{2.25in}
\begin{document}
\vspace*{0.2in}
\begin{flushleft}
{\Large
\textbf\newline{Detecting drug-drug interactions using artificial neural networks and classic graph similarity measures} % Please use "sentence case" for title and headings (capitalize only the first word in a title (or heading), the first word in a subtitle (or subheading), and any proper nouns).
}
\newline
\\
Guy Shtar,\textsuperscript{1*} %\Yinyang
Lior Rokach,\textsuperscript{1}
Bracha Shapira,\textsuperscript{1}
\\
\bigskip
\textbf{1} Department of Software and Information Systems Engineering, Ben-Gurion University of the Negev, Beer-Sheva, Israel
\\
\bigskip
* Shtar@post.bgu.ac.il
\end{flushleft}
\section*{Abstract}
Drug-drug interactions are preventable causes of medical injuries and often result in doctor and emergency room visits. Computational techniques can be used to predict potential drug-drug interactions. We approach the drug-drug interaction prediction problem as a link prediction problem and present two novel methods for drug-drug interaction prediction based on artificial neural networks and factor propagation over graph nodes: adjacency matrix factorization (AMF) and adjacency matrix factorization with propagation (AMFP). We conduct a retrospective analysis by training our models on a previous release of the DrugBank database with 1,141 drugs and 45,296 drug-drug interactions and evaluate the results on a later version of DrugBank with 1,440 drugs and 248,146 drug-drug interactions. Additionally, we perform a holdout analysis using DrugBank. We report an area under the receiver operating characteristic curve score of 0.807 and 0.990 for the retrospective and holdout analyses respectively. Finally, we create an ensemble-based classifier using AMF, AMFP, and existing link prediction methods and obtain an area under the receiver operating characteristic curve of 0.814 and 0.991 for the retrospective and the holdout analyses. We demonstrate that AMF and AMFP provide state of the art results compared to existing methods and that the ensemble-based classifier improves the performance by combining various predictors. Additionally, we compare our methods with multi-source data-based predictors using cross-validation. In the multi-source data comparison, our methods outperform various ensembles created using 29 different predictors based on several data sources. These results suggest that AMF, AMFP, and the proposed ensemble-based classifier can provide important information during drug development and regarding drug prescription given only partial or noisy data. Additionally, the results indicate that the interaction network (known DDIs) is the most useful data source for identifying potential DDIs and that our methods take advantage of it better than the other methods investigated. The methods we present can also be used to solve other link prediction problems. Drug embeddings (compressed representations) created when training our models using the interaction network have been made public.
%\linenumbers

\section*{Introduction}
Adverse drug events are often preventable causes of medical injuries, and adverse drug reactions (ADRs) are estimated to be the fourth leading cause of death in the U.S., ahead of pulmonary disease, diabetes, AIDS, pneumonia, accidents, and automobile fatalities~\cite{FDA_drug_interactions}. The cost attributed to ADRs is estimated to be over \$1,000 per patient per year in the US~\cite{doi:10.1001/jama.1997.03540280045032}. Estimates of the number of patients harmed due to drug interactions range from 3-5\% of all medication errors within hospitals. Additionally, drug interactions are the cause of many patient visits to physicians and emergency units~\cite{pmid10192758, pmid17047216}. Thirty-six percent of older adults in the U.S. regularly use five or more medications or supplements, and 15\% are potentially at risk for a major drug-drug interaction (DDI)~\cite{pmid26998708}. The American Geriatrics Society has identified the consideration of drug-disease and drug-drug interactions as a key element of optimal care for older adults with multimorbidity~\cite{doi:10.1111/j.1532-5415.2012.04188.x}.
DDI prediction during the clinical experiments conducted in order to approve a new drug is difficult~\cite{CORRIGAN2002497}. Clinical trials for new drugs don’t address the issue of DDI directly, and potential DDIs are often not discovered until the third phase of a clinical trial or once the drug is already on the market. The most practical way to explore the large number of drug combinations for detecting interacting drugs is through in silico drug-drug interaction detection, and in this paper, we propose a computational method for DDI detection. \par 
In recent years, the detection of potential DDIs using computational techniques has gained attention; previous research has used techniques based on drug-drug interaction similarities~\cite{10.1371/journal.pone.0058321}, side effect similarities~\cite{Zhang2015}, structural similarities~\cite{RyuE4304}, or a combination of
various similarity measures~\cite{Gottlieb592, Zhang2017, 10.1371/journal.pone.0140816}. Other works use natural language processing (NLP) techniques to train word embedding using document collections such as PubMed, PMC, MEDLINE, and Wikipedia; the embeddings are later used to predict DDIs~\cite{10.1371/journal.pone.0190926}. Computational methods often require a large amount of data for optimization. For example, when evaluating a new drug using structural-based similarity methods, the method will require data showing a strong, well established history for structurally similar drugs in order to accurately detect drug interactions. Side effect similarity-based methods require data for drugs with similar side effects, etc. We compare our proposed methods with other methods which were created using various data types, and the results indicate that the interaction network (known DDIs) is the most useful data source for identifying potential DDIs. Like other data sources, the interaction network has limitations. For example, when evaluating a new drug with no known interactions, the DDI network will not be helpful. Therefore, the drug-drug interaction prediction problem should be investigated using various data types. \par
\begin{figure}[H]

\includegraphics[width=\textwidth]{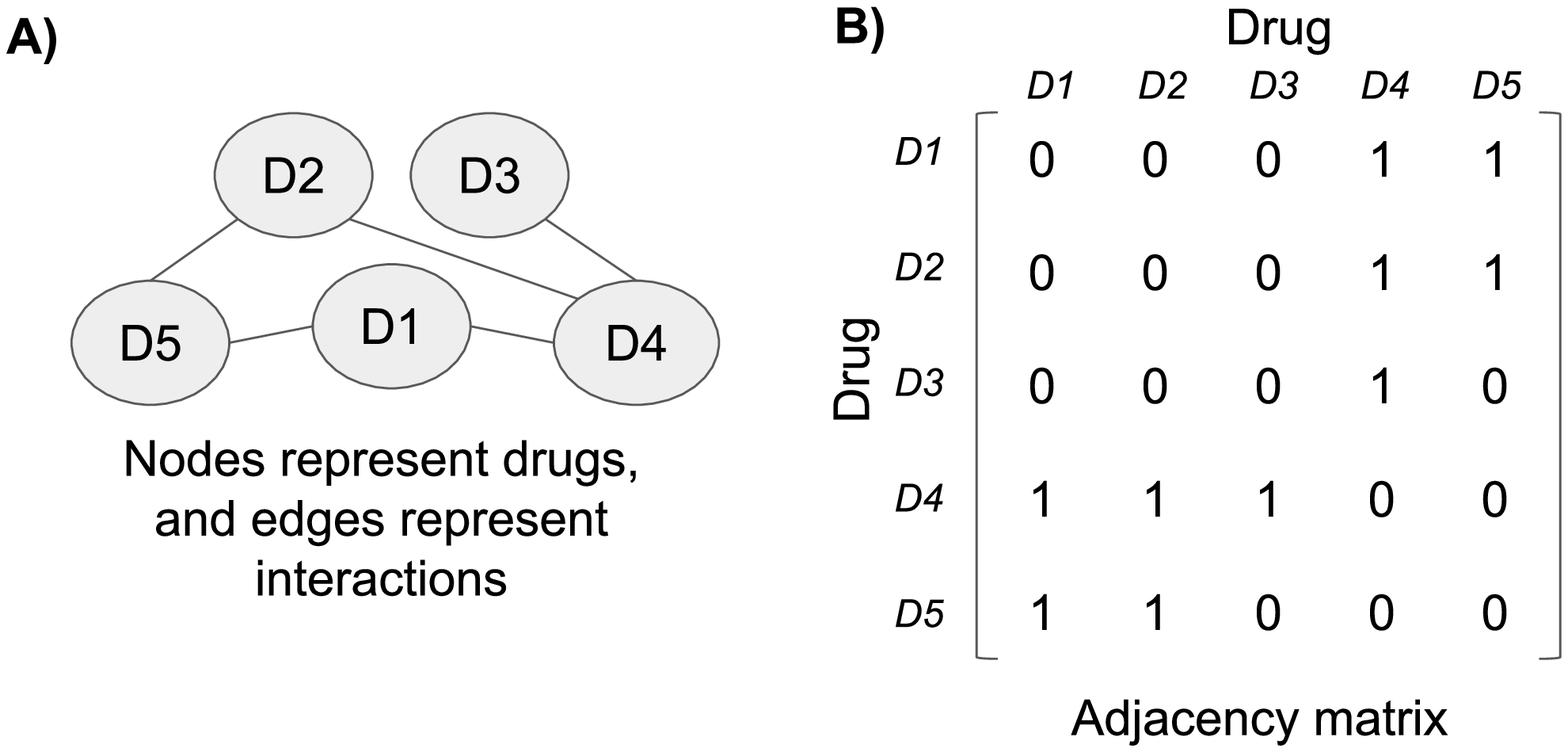}
\caption{{\bf Tackling the DDI prediction problem as a link prediction problem.} \textbf{A)} A DDI graph is created: nodes represent drugs, and edges represent interactions. \textbf{B)} The DDI graph is represented by an adjacency matrix, rows and columns represent drugs, and a value of one in the matrix indicates an existing interaction; for example, the cell in the first row and the last column represents the interaction between D1 and D5. In a link prediction problem, a score is calculated to every non-existing interaction.}
\label{fig0}
\end{figure}
\begin{figure}[H]

\includegraphics[width=\textwidth]{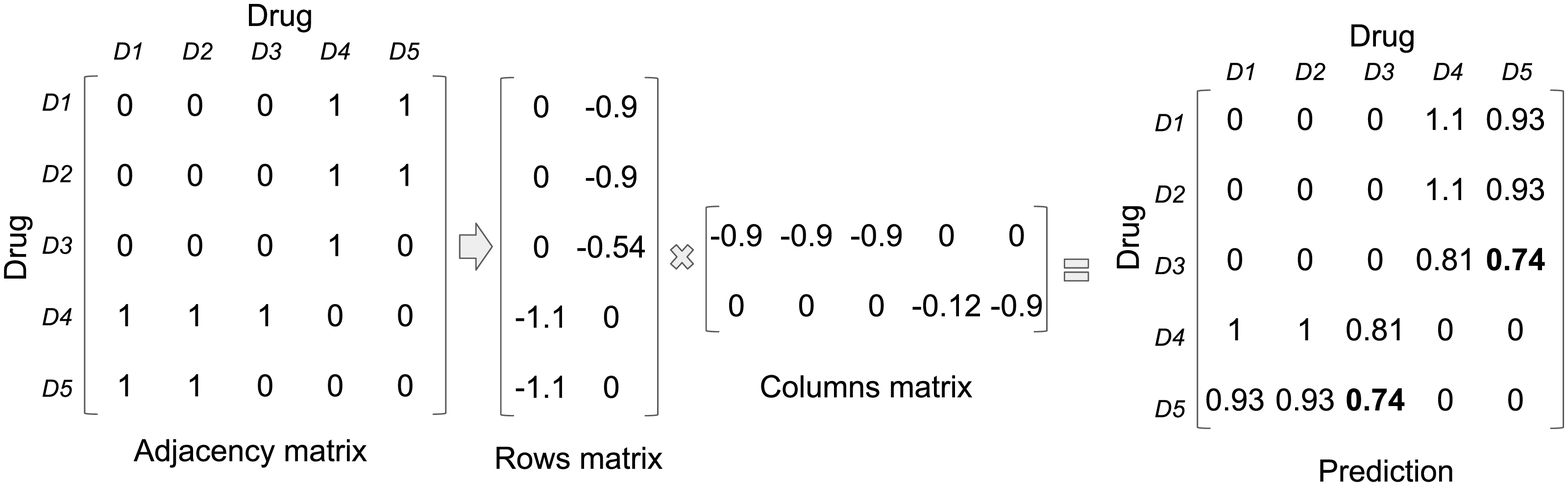}
\caption{{\bf Link prediction using matrix factorization for DDI prediction.} The dimension of the adjacency matrix is reduced by factorizing it into two lower ranked matrices. By multiplying the matrices a score is calculated for every existing and non-existing interaction. In this case, an interaction between D5 and D3 is very likely to exist. A score is also given to existing links: the interaction between D1 and D4 is stronger than the interaction between D1 and D5.}

\label{fig01}
\end{figure}

DDI detection can be seen as a special case of link prediction in a graph. In a link prediction problem, we seek to accurately predict the edges (interactions) between nodes (drugs) that will be added to the network. We approach the DDI prediction problem as a link prediction problem. 
Perhaps the most basic approach is to rank edges based on the idea that two nodes $x$ and $y$ are more likely to form a link if their sets of neighbors have a large overlap; this follows the natural intuition that such nodes $x$ and $y$ represent drugs with many interacting drugs in common, and hence are more likely to interact. 
Matrix factorization is another approach for resolving link prediction problems. Matrix factorization (MF) is the factorization of a matrix into a product of matrices; this technique is widely used for dimensionality reduction, specifically in the field of recommender systems. In recent years, successful attempts have been made to factorize a matrix using deep neural networks~\cite{He:2017:NCF:3038912.3052569, WU201846, FAN201834}. Figures~\ref{fig0} and~\ref{fig01} demonstrate how the DDI prediction problem can be solved by factorizing and reducing the dimensionality of the adjacency matrix representing the drug-drug interaction graph. For clarity, in figure~\ref{fig01} the prediction matrix is made symmetric by averaging opposite cells. The figures also demonstrate the drawbacks of matrix factorization addressed by this research: first, the decomposition is not symmetric. Since the row vectors and column vectors of the adjacency matrix are identical, the transpose of the columns matrix should be equal to the rows matrix. The second drawback is that the score is not bound, it can be limited to the range $[0,1]$.

In this paper, we introduce AMF and AMFP, two novel methods for predicting DDIs based on artificial neural networks and the implementation of factor propagation over the interaction network. Unlike some of the methods presented in previous research, AMF and AMFP use only known drug interactions as input and predict currently unknown drug interactions. Additionally, most of the previous studies based their work on similarity measures, but AMF and AMFP are based on machine learning techniques, specifically on neural networks. We compare AMF and AMFP to existing methods which are based on multiple data sources and create an ensemble-based classifier using AMF, AMFP, and other well-known link prediction methods. Compared with existing methods, our methods produce better performances using only known DDIs as the data source, and the statistical analysis demonstrates that the performance improvements achieved by our method are statistically significant.
In this paper, we make three key contributions: (1) we formulate a new artificial neural network-based method for link prediction, (2) we demonstrate its effectiveness for the drug-drug interaction prediction challenge, conducting extensive evaluations with real data to show the superiority of the interaction network (known DDIs) as a data source; to show the superiority of our method, we create drug embeddings for all available drugs, and (3) we create an ensemble-based classifier to demonstrate the benefit of combining existing high-performing classifiers. The preprocessing, methods, and drug embeddings developed, calculated, and used in this research were implemented and have been made public: \url{https://github.com/goolig/DDI_prediction}.

\section*{Materials and methods}
\subsection*{Problem formulation}
We approach the drug-drug interaction prediction problem as a link prediction problem.
Suppose we have an undirected drug interaction network $G = (V, E)$ in which each edge $e = (u, i) ∈ E$ represents an interaction between drugs $u$ and $i$. Note that throughout the paper we use the terms graph node and drug interchangeably. We use two versions of the drug interactions graph: $G$ and $G'$.
For two time snapshots $t < t'$ let G denote the graph constructed using the known interactions at time $t$, and $G'$ denote the constructed graph using the known interactions at time $t'$. This is a concrete formulation of the drug-drug interaction prediction problem: we give an algorithm access to network $G$, and it must then output a list of edges not present in $G$, which are predicted to appear in network $G'$. We refer to $t$ as the training release date and $t'$ as the test release date. Of course, drug interaction networks grow through the addition of nodes (drugs) as well as edges. The training process uses only existing interactions to predict unknown ones. It is not sensible to seek predictions for edges whose endpoints are not present in the training interval, as such a prediction will be based on partial information in terms of link prediction. We use the adjacency matrix to represent $G$ and $G'$, and the matrices are used to train AMF and AMFP:
Let M denote the number of drugs. We define the drug-drug interaction matrix $Y \in ℝ^{MXM}$ as follows:

\begin{equation}
  y_{i,j} = \left\{
\begin{array}{ll}
      1, & \textrm{if interaction exits between drugs i,j;}  \\ 0, & \textrm{otherwise.} \\      
\end{array} 
\right. \end{equation}
Here, a value of one for $y_{i,j}$ indicates an existing interaction between drugs $i$ and $j$, however a value of zero does not mean that an interaction does not exist - it could be that the interaction has not yet been discovered.

\subsection*{Adjacency matrix factorization}
\begin{figure}[H]
\includegraphics[width=\textwidth]{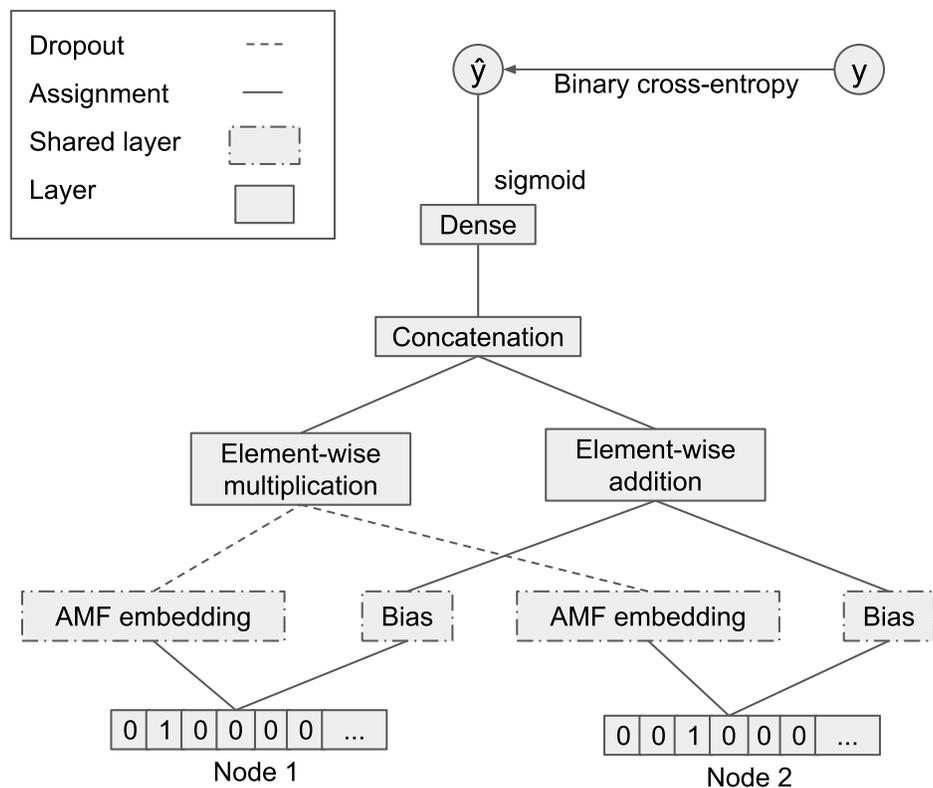}
\caption{{\bf Overview of AMF's architecture.}
Drugs are represented as nodes; embedding layers (which act as latent factors) and biases are shared between input nodes. Dropout is used as a regularization mechanism for preventing overfitting.}
\label{fig1}
\end{figure}

Traditionally, matrix factorization associates each row element $i$ and column element $j$ with a corresponding latent vector $p_i$ and $q_j$. The estimate of the corresponding cell $y_{i,j}$ of the matrix is given by the inner product of the vectors:
\begin{equation}
\hat{y}_{i,j} = \sum_{w=1}^kp_{iw}q_{jw},
\end{equation}
where the vectors $p_i$ and $q_j$ are the latent factors, sometimes also referred to as representations, because they can be used as an alternative representation of the original row and column objects. The space size $k$ is a parameter, usually set to a much lower value than the original space size. Using an extremely low $k$ value might lead to underfitting; on the other hand, extremely high values might lead to overfitting. 
MF is sometimes improved by using a bias value corresponding to the row and column elements: 
\begin{equation}
\hat{y}_{i,j} = \mu + b_i + b_j + \sum_{w=1}^kp_{iw}q_{jw},
\end{equation}
where $\mu$ is the average value over the whole matrix, and $b_i$ and $b_j$ are the bias values for the row and column elements, correspondingly. The parameters are typically learned using optimization techniques such as stochastic gradient decent. Regularization techniques are often used during the optimization process.
AMF (adjacency matrix factorization) performs matrix factorization on the adjacency matrix of $G$. Because the graph $G$ is undirected, the adjacency matrix $M$ is symmetric. Therefore, it is sufficient to use a single vector and bias value shared between the rows and columns to estimate $M$'s cells.
% two steps: (1)	Optimization:
% and a factor propagation technique which allows sharing the	characteristics of interacting drugs
To allow precise DDI prediction, we use an artificial neural network-based model that encompasses the linear structure of the interaction network. The method is based on optimizing the latent factors of each drug in the network; the latent factor is an $k$-dimensional vector.
Figure~\ref{fig1} provides an overview of AMF’s architecture. AMF takes a one-hot encoding representation of the two nodes under consideration as input. The output of AMF is binary; one indicates an existing interaction, and zero indicates no interaction between the two input nodes. Note that only an existing drug interaction network is required to train the proposed neural network. No other domain-specific information is required.

The use of a simple inner product to estimate complex drug-drug interactions in the	low-dimensional latent space might be oversimplistic. The complexity of this model can be increased by increasing the embedding layer size, however that might cause overfitting. The space size $k$ should be carefully tuned during the training phase. Matrix factorization and AMF are closely related to singular value decomposition (SVD). AMF's main improvements in the optimization stage compared to matrix factorization and SVD are: (1) sharing the latent vectors of the rows and columns, and (2) optimizing the weights of the element-wise multiplication and the bias, rather than just using a dot product and adding the biases.	
AMF is optimized using the adaptive moment estimation (Adam) optimization algorithm~\cite{ADAM_2014} which adjusts the learning rate for each parameter using estimates of the first and second moments of the gradients. MF and SVD are aimed at minimizing the mean square error; for a classification task such as a link prediction problem, binary cross-entropy is more appropriate. Hence, the loss function used in AMF is binary cross-entropy, defined as follows:

\begin{equation}
L = \sum_{i,j\in{Y}} -y_{i,j}\log\hat{y}_{i,j} - (1-y_{i,j})\log{(1-\hat{y}_{i,j})},
\end{equation}
where $Y$ is the set of instances (drug pairs), $y_{i,j}$ is the true label which represents the existence or absence of an interaction, and $\hat{y}_{i,j}$ is the predicted value. $Y$ is created using negative sampling, where all positive samples are used (existing drug interactions), and a relative number of negative samples are drawn randomly in each epoch; the number of negative samples to be sampled is a hyperparameter of the model that should be tuned. In this research we sample one negative sample for each given positive sample.
\subsection*{Adjacency matrix factorization with propagation}
AMF excels in the holdout analysis where data is randomly sampled for the testing set (see \nameref{Results} section for more details). Nevertheless, despite all of the regularization techniques applied to it, AMF's generalization ability and performance are poor in the retrospective analysis where a new unseen version of the dataset is used. The underlying mechanism that causes an interaction to be discovered and added to the database is not random - it depends on mediators such as new interactions discovered between substances contained in the drugs, the drug's prevalence, and more. To perform well in the retrospective analysis, higher generalization ability is required from the model.
Adjacency matrix factorization with propagation (AMFP) is an extension of AMF. In AMFP, the same model used in AMF is used, but an additional step is performed: propagating the latent factors of each drug to its interacting drugs; latent factor propagation is controlled by a propagation factor, which controls the weight of the original latent factor (which was optimized in the previous step) compared to the weight of the latent factor of the interacting drugs.
Algorithm~\ref{propgateFactors} describes the propagation procedure. Each node’s latent factor is shared with the node’s neighborhood. The parameter $\alpha$ is the propagation factor, which controls how much information will be passed from the neighboring nodes. The value of $\alpha$ should be optimized during the training process.

\begin{algorithm}
    \caption{Latent factor propagation}
    \label{propgateFactors}
    \begin{algorithmic}[1] % The number tells where the line numbering should start    
        \Procedure{propagate\_factors}{Graph G = (V, E), Latent factors P, Propagation factor $\alpha$}
            \State $P'\gets empty \ list$ // holds the new latent factors 
            \For{vertex $v1$ in $G$} 
            	\State $Q \gets empty \ latent \ factor$
                \For{vertex $v2$ in $\Gamma(v1)$} 
                \State $Q \gets Q + \frac{1}{|\Gamma(v1)|} \cdot P_{v2}$
                \EndFor
                \State $P'_{v1} \gets \alpha \cdot Q + (1-\alpha) \cdot P_{v1} $
            \EndFor\label{euclidendwhile}
            \State \textbf{return} $P'$
            %\Comment{The gcd is b}
        \EndProcedure
    \end{algorithmic}
\end{algorithm}

Given a node $v ∈ V$; the neighborhood of $v$ is defined by $\Gamma(v)$ and represented by the set of $v$’s interacting drugs. The lists $P$ and $P'$ contain the original latent factors (embeddings) and the latent factors resulting from the propagation process correspondingly. Each list contains vectors (each vector has $k$ elements) representing the latent factor of the nodes. $\alpha$ is the propagation factor; when $\alpha$ reaches a value of one, the original latent factor of each node is discarded, and a new latent factor is created based on the neighborhood of the node. On the other extreme, when $\alpha$ reaches a value of zero, the latent factors created in the previous step are used, and the propagation step does not change the model. When $\alpha$'s value is equal to zero, the results of AMPF and AMF are equivalent. Propagating the factors is expected to improve the generalization ability of the model by combining the factors of interacting drugs. This logic follows the assumption that interacting drugs share some common characteristics.

\subsection*{Link prediction similarity measures}\label{sssec:lpsm}
Link prediction similarity measures can be viewed as pre-engineered features which leverage domain knowledge for link prediction in graphs. This subsection is devoted to formulating and explaining the motivation behind the similarity measures used in this research for creating an ensemble-based classifier and evaluation. \par
The common neighbors between two given nodes $u, v ∈ V$ refers to the size of the set of common neighbors that both $u$ and $v$ possess. The formal common neighbors definition is:
\begin{equation}
S^{CN}(u,v) = |\Gamma(v) ∩ \Gamma(u)|.
\end{equation}
The relevance of the common neighbors feature is very intuitive. It is expected that the larger the size of the common neighborhood, the higher the chances are that both vertices will be connected. The common neighbors feature has been widely used in past work on link prediction on several datasets and was found to be very helpful~\cite{6033365}. Using the common neighbors measure, we formulate the average common neighbors for two nodes: 
\begin{equation}
\widetilde{S}^{ACN}(u,v) = \frac{1}{|\Gamma(v)|}\sum_{w∈ \Gamma(v)}{S^{CN}(w,u)}.
\end{equation}
The average common neighbors measure provided above is not symmetric, due to the normalizing factor ${|\Gamma(v)|}$; we formulate it as a symmetric measure by averaging its two possible values for a pair of nodes:
\begin{equation}
S^{ACN}(u,v) = \frac{\widetilde{S}^{ACN}(u,v) + \widetilde{S}^{ACN}(v,u) }{2}.
\end{equation}
The Jaccard coefficient is a well-known similarity measure, widely used for link prediction~\cite{6033365}. For two nodes $u, v ∈ V$ the Jaccard coefficient is defined as follows:
\begin{equation}
S^{Jaccard}(u,v) = \frac{|\Gamma(v) ∩ \Gamma(u)|}{|\Gamma(v) \cup	 \Gamma(u)|}.
\end{equation}
As with common neighbors, we formulate the average Jaccard coefficient for two vertices as follows:
\begin{equation}
\widetilde{S}^{AJ}(u,v) = \frac{1}{|\Gamma(v)|}\sum_{w∈ \Gamma(v)}{S^{Jaccard}(w,u)},
\end{equation}
and its symmetric version is given as follows:
\begin{equation}
S^{AJ}(u,v) = \frac{\widetilde{S}^{AJ}(u,v) + \widetilde{S}^{AJ}(v,u) }{2}.
\end{equation}
The Adamic/Adar index~\cite{Adamic2003} is a similarity measure used to predict links in social networks:
\begin{equation}
S^{AA}(u,v) = \sum_{w∈ \Gamma(v) \cap \Gamma(u)} \frac{1}{\log{|\Gamma(w)|}}.
\end{equation}
Lastly, we present ${Katz}_b$ which exponentially sums the number of shortest paths of different lengths between two nodes: 
\begin{equation}
{Katz}_b(u,v) = {\Sigma}_{\ell=1}^{\infty}\beta^{\ell}\cdot{|paths|}_{u,v}^{<\ell>},
\end{equation}
where ${|paths|}_{u,v}^{<\ell>}$ is the number of paths between $u$ and $v$ of length $\ell$, and $\beta$ is a parameter controlling the weight given to shorter paths compared to the weight given to longer ones. In practice, a truncated Katz measure is usually used:
\begin{equation}
{Katz}_b(u,v) = {\Sigma}_{\ell=1}^{b}\beta^{\ell}\cdot{|paths|}_{u,v}^{<\ell>}.
\end{equation}
In this research, we use $b=3$ due to resource limitations. ${Katz}_b$ was found to be very helpful for link prediction in previous works~\cite{4118528}.
The final list of link prediction similarity measures used in this research consists of: average common neighbors, average Jaccard coefficient, Adamic/Adar, and ${Katz}_b$. The similarity measures between two nodes used by Fire et al.~\cite{Fire:2014:CEL:2542182.2542192}, such as the shortest path length between nodes, cosine distance between nodes, and dividing the graph into communities and then comparing two nodes' communities were tested and discarded due to poor performance.

\subsection*{Creating an ensemble-based classifier}
Ensembles are meta-algorithms used to combine various classifiers. They can reduce the variance and bias of the base models and improve the predictions in general. One such ensemble method is XGBoost~\cite{Chen:2016:XST:2939672.2939785} which achieved state of the art results in multiple tasks and competitions. XGBoost employs gradient boosting where models are created stage wise using weak predictors, usually using prediction trees. In each stage, the model seeks to improve the performance of the model created in the previous stage.
We train an ensemble classifier using XGBoost, based on the link prediction similarity measures presented above: average common neighbors, average Jaccard coefficient, Adamic/Adar, and ${Katz}_b$. Additionally AMF or AMFP (the method that performs better) and the method proposed by Vilar et al.~\cite{10.1371/journal.pone.0058321} are fed to the ensemble classifier. This meta-algorithm can be easily extended to include additional features.

\section*{Evaluation} \label{Evaluation}
In this section, we present experiments with the aim of evaluating AMF, AMFP, and the ensemble-based classifier. Our evaluation is based on two evaluation schemes: a retrospective analysis using approved drugs from three versions of the DrugBank database~\cite{pmid29126136} and a holdout analysis using a current version of the database. We use state of the art benchmarks.

\begin{figure}[H]

\includegraphics[width=\textwidth]{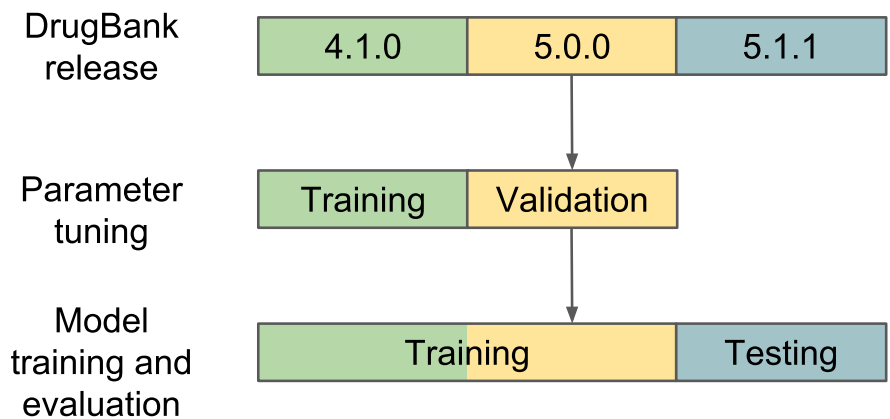}
\caption{{\bf Retrospective evaluation scheme.} Parameter tuning is performed using DrugBank release 4.1.0 and 5.0.0. The previous release is used to train the model, and the latter is used to validate the results. The final model is trained using the parameters obtained in the validation stage with the data from release 5.0.0 (which contains the data from release 4.1.0 with some additions and changes) and tested using release 5.1.1.}

\label{fig2}
\end{figure}

Figure~\ref{fig2} illustrates the validation and testing scheme for the retrospective analysis using three versions of the DrugBank database. Major changes were made between the versions - specifically, a large number of interactions were added to the more recent version. For the validation process, we aligned versions 4.1.0 and 5.0.0 by only using drugs which appear in both versions. The same was done for versions 5.0.0 and 5.1.1 when training and testing the final model. Version 4.1.0 from December 2014 contains 11,284 interactions, and versions 5.0.0 from June 2016 and 5.1.1 from July 2018 contain 45,296 and 248,146 interactions respectively. Versions 4.1.0, 5.0.0, and 5.1.1 respectively contain 1,141, 1,440, and 2,149 drugs. To test whether our model could predict pharmacodynamic as well as pharmacokinetic  interactions, we adopt a similar evaluation scheme to the one used by Vilar et al.~\cite{10.1371/journal.pone.0058321}. We use DrugBank annotations to identify any interactions between drugs with shared metabolism by a cytochrome p450 (CYP) metabolizing enzyme (1A2, 2B6, 2C8, 2C9, 2C19, 2D6, 2E1, 3A4, 3A5 and 3A7). Such interactions are removed from the test set (release 5.1.1), and the rest of the retrospective analysis is executed normally. 

In addition to the retrospective analysis, we perform a holdout evaluation using release 5.1.1, the latest release available during this research. The setup of the holdout evaluation is as follows: 30\% of randomly selected existing and non-existing interactions are used as a test set, and the rest of the data is used as a training set, 10\% of the data is used for validation (parameter tuning) during training. For both evaluation techniques, the model is retrained after the validation process, with the combined training and validation data, using the tuned parameters. 
In a holdout evaluation, the interactions are randomly selected, while in reality some interactions are more likely to be found earlier depending on the popularity of the drug, the prevalence of the interaction, etc. For these reasons, the retrospective analysis is a stronger evaluation scheme, however we perform a holdout evaluation for comprehensiveness and to comply with previous research.

\subsection*{Metrics}
The primary evaluation metric we use is the area under the ROC curve (AUROC). We also assess the area under the  precision-recall curve (AUPR), because it was argued to be relevant for link prediction problems~\cite{Yang2015}. We plot the ROC curve and the average precision @ n, where the precision of each drug's prediction is averaged at different values of $n$. Lastly, we plot precision @ n which evaluates the top $n$ most confident predictions of the model. We acknowledge the importance of precision over recall in the DDI problem, therefore we plot the two precision graphs in addition to the other metrics.

\subsection*{Baselines}
We compare our proposed method with the following methods:
\begin{itemize}
\item The method suggested by Vilar et al.~\cite{10.1371/journal.pone.0058321}. This method is based on drug interaction profile fingerprints (IPFs). The model uses IPFs to measure the similarity of pairs of drugs and generates new putative drug-drug interactions from the non-intersecting interactions of a pair. Their method uses the same input data used by the method proposed in this paper.
\item The link prediction similarity measures presented earlier: average common neighbors, average Jaccard coefficient, Adamic/Adar, and ${Katz}_b$.
\item An XGBoost model trained using all of the models described in the previous bullets and AMF or AMFP (we use the model which performs better). This model is used to demonstrate the power of combining several strong methods, rather than being used in comparison with the other methods mentioned above, as a comparison between a regular model and an ensemble is inappropriate.
\item The methods used by Zhang et al.~\cite{Zhang2017}. The following multi-source data is used: substructure data, drug target data, drug enzyme data, drug transporter data, drug pathway data, drug indication data, drug side effect data, drug side effect data, and known drug-drug interactions. The neighbor recommender method and random walk method are used to create DDI prediction models. Using 29 prediction models, including 28 similarity-based models and one perturbation matrix model, three ensembles are created based on logistic regression with L1 and L2 regularization and a genetic algorithm. The ensembles are compared to two existing methods presented by Vilar et al.~\cite{10.1371/journal.pone.0058321, 10.1136/amiajnl-2012-000935} and three methods presented by Zhang et al.~\cite{Zhang2015}, our methods are compared indirectly with these methods by using this baseline and the same dataset. The authors include data from different sources, creating a diverse dataset. Unfortunately, the dataset is available for just a single point in time, which does not allow a retrospective analysis. To compare the methods presented by Zhang et al.~\cite{Zhang2017} to ours, we adopt the cross-validation scheme used in the original research. We use three and five-fold cross-validation, repeat each experiment five times and use pairwise t-test on the results.

\end{itemize}

We implemented AMF and AMFP using Keras~\cite{chollet2015keras}. The method suggested by Vilar et al. was implemented in Python; we used the original implementation and data for the methods suggested by Zhang et al. For Adamic/Adar and several other methods we used the NetworkX implementation~\cite{osti_960616}. Other methods, such as ${Katz}_b$, were implemented in Python. For XGBoost, the implementation proposed by the authors was used.
We assess the AUROC score of AMF, AMFP, and each of the baselines for significance with a paired test using the algorithm described by Sun and Xu~\cite{6851192}.

\subsection*{Parameter tuning}
To determine the hyperparameters of AMF and AMFP, we used the procedure described above. All weights are randomly initialized using the Glorot normal initializer~\cite{pmlr-v9-glorot10a}. The following batch sizes were used: 128, 256, 512, and 1024, and learning rates in the range of 0.1 - 0.0001 were tested. We evaluate the following number of factors (embedding sizes): 32, 64, 128, 256, 512, and 1024, dropout levels in the range of 0-0.9, the number of epochs in the range of 1-50, and propagation factors in the range of 0.0-1.0. For XGBoost, the parameters were optimized using randomized grid search, where combinations of parameters were drawn randomly from a given list and evaluated.

\section*{Results} \label{Results}
In this section, we report the results of AMF, AMFP, and the baselines using the evaluation techniques described in the previous section.

\subsection*{Holdout analysis}
\begin{figure}[H]
\includegraphics[]{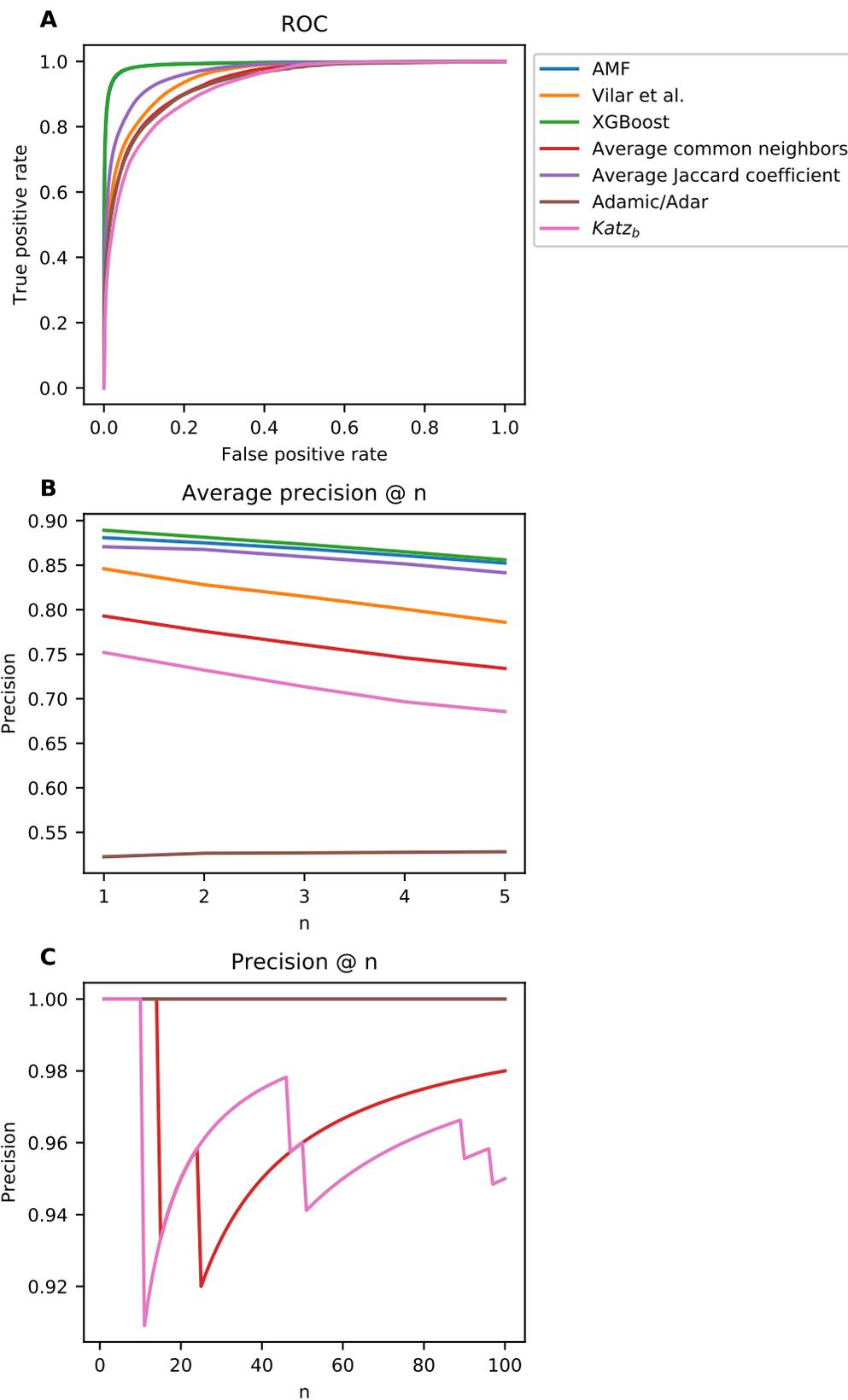} 
\caption{{\bf Holdout analysis results.}
\textbf{A)} Receiver operating characteristic curves; \textbf{B)} Per drug average precision @ n; \textbf{C)} Precision @ n.}

\label{fig3}
\end{figure}

Holdout analysis is performed by using 70\% of the data in DrugBank release 5.1.1 to train the models; the rest of the existing and non-existing interactions are used for evaluation. Fig~\ref{fig3}A shows the ROC curve for AMF and each of the baselines. Table~\ref{table1} presents the AUROC and AUPR values for each model. The AUROC of each pair of models was tested for significance; we report a $p-value<10^{-4}$ for all tests. Figure~\ref{fig3}B shows the average precision @ n (per drug), where $n$ ranges from one to five, and figure~\ref{fig3}C shows the precision @ n, where $n$ ranges from one to 100. The optimal value for AMFP's $\alpha$ is zero, hence its performance is equivalent to that of AMF. Therefore, we do not present its results in the holdout analysis, and it is not used in the XGBoost model trained using the holdout data.

% Place tables after the first paragraph in which they are cited.
\begin{table}[!ht]
\begin{adjustwidth}{-2.25in}{0in} % Comment out/remove adjustwidth environment if table fits in text column.
\centering
\caption{
{\bf Area under the ROC and precision-recall curves for the holdout analysis.}}
\begin{tabular}{|l+l|l|l|}
\hline
\multicolumn{1}{|l|}{\bf Algorithm} & \multicolumn{1}{|l|}{\bf AUROC curve} & \multicolumn{1}{|l|}{\bf AUPR curve}\\ \thickhline
$XGBoost$ & 0.991 & 0.960 \\ \hline
$AMF$ & 0.990 & 0.950 \\ \hline
$Vilar\:et\:al.$~\cite{10.1371/journal.pone.0058321} & 0.952 & 0.784 \\ \hline
$Average\:common\:neighbors$ & 0.938 & 0.738 \\ \hline
$Average\:Jaccard\:coefficient$ & 0.967 & 0.840 \\ \hline
$Adamic/Adar$ & 0.933 & 0.728 \\ \hline
${Katz}_b$ & 0.924 & 0.675\\ \hline
\end{tabular}
\begin{flushleft}
\end{flushleft}
\label{table1}
\end{adjustwidth}
\end{table}

\subsection*{Retrospective analysis}
\begin{figure}[H]
\includegraphics[]{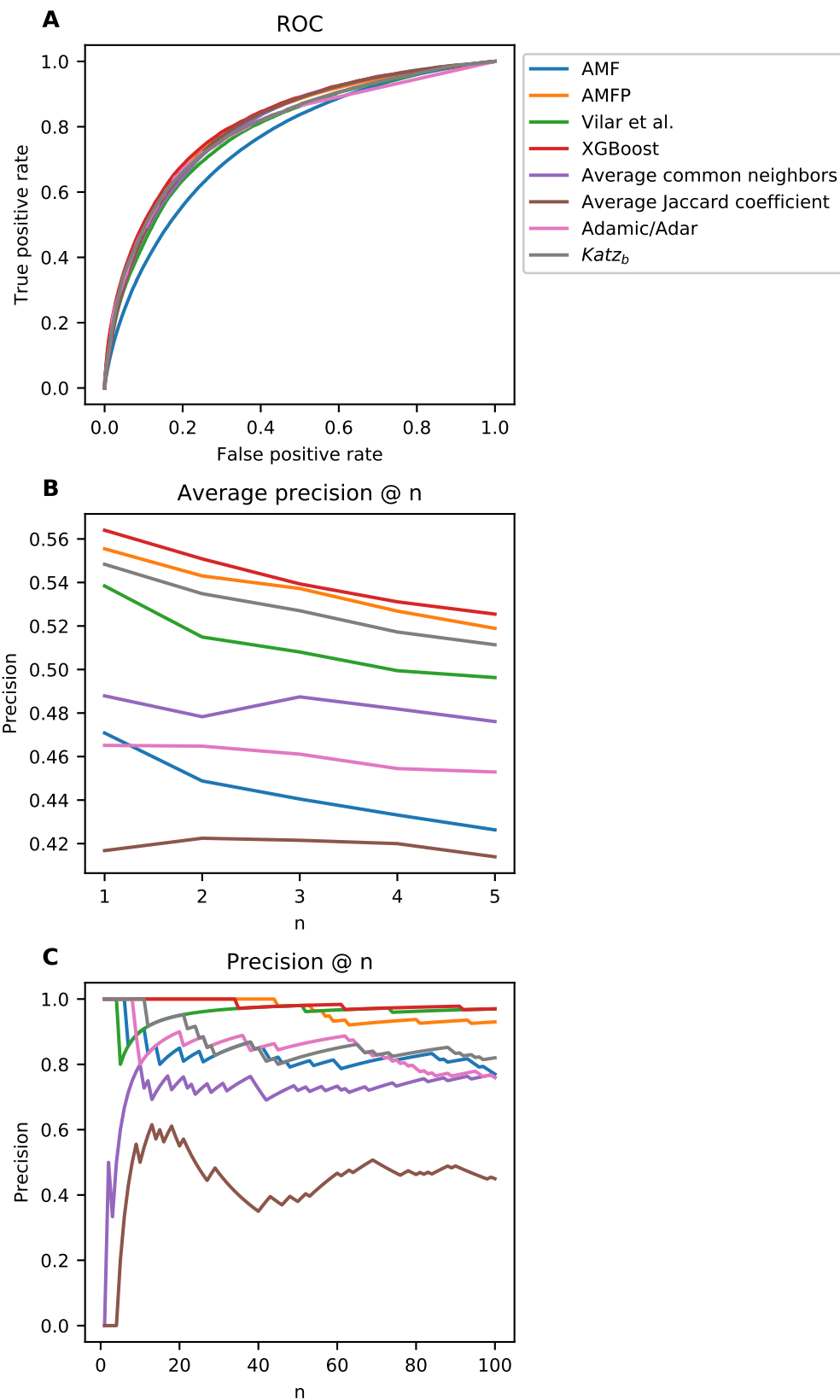}
\caption{{\bf Retrospective analysis results.}
\textbf{A)} Receiver operating characteristic curves; \textbf{B)} Per-drug average precision @ n; \textbf{C)} Precision @ n.}

\label{fig4}
\end{figure}

Retrospective analysis is performed by training the models on an older version of DrugBank and evaluating the models using a more recent version of DrugBank.
Fig~\ref{fig4}A shows the ROC curve for AMF, AMFP, and each of the baselines. Table~\ref{table2} presents the AUROC and AUPR values for each model. The AUROC of each pair of models was tested for significance; we report a $p-value<10^{-4}$ for all tests. Figure~\ref{fig4}B shows the average precision @ n (per drug), where $n$ ranges from one to five, for the first interaction ($n=1$), the accuracy is about 56\% for both AMFP and the XGBoost ensemble. Figure~\ref{fig4}C shows the precision @ n, where $n$ ranges from one to 100. XGBoost was trained using AMFP's predictions and without AMF's predictions because of their superiority. XGBoost has the best performance in terms of the AUROC and AUPR curves, followed by AMFP. The average precision @ n and precision @ n graphs demonstrate the XGBoost model superiority, it performs best for almost all values of n.
To test whether our model could predict pharmacodynamic as well as pharmacokinetic interactions, we removed any interactions between drugs with shared metabolism by a cytochrome p450 (see \nameref{Evaluation} section). A total of 56,874 interactions were removed (37.7\% of the interactions). We report an AUROC of 0.775 for AMFP and 0.705 for AMP, a performance reduction of 0.032 and 0.043 respectively. For reference, the performance reduction is 0.044 for the method developed by Vilar et al.~\cite{10.1371/journal.pone.0058321}. These results suggests that AMF and AMFP take different pharmacological effects caused by pharmacokinetic and pharmacodynamic characteristics of the drugs into account.
While the relative performance between the retrospective analysis and the holdout analysis is somewhat similar for all models, the absolute differences are obvious. This phenomena can be explained by the fact that each DrugBank release is a closed system of interactions known at a given time, sometimes derived from interactions between substances contained in different drugs. The absolute differences between results demonstrate the weakness of holdout evaluation and its difficulty in simulating real-world scenarios compared to retrospective evaluation.

% Place tables after the first paragraph in which they are cited.
\begin{table}[!ht]
\begin{adjustwidth}{-2.25in}{0in} % Comment out/remove adjustwidth environment if table fits in text column.
\centering
\caption{
{\bf Area under the ROC and precision-recall curves for retrospective analysis.}}
\begin{tabular}{|l+l|l|l|}
\hline
\multicolumn{1}{|l|}{\bf Algorithm} & \multicolumn{1}{|l|}{\bf AUROC curve} & \multicolumn{1}{|l|}{\bf AUPR curve}\\ \thickhline
$XGBoost$ & 0.814 & 0.425 \\ \hline
$AMFP$ & 0.807 & 0.417 \\ \hline
$AMF$ & 0.748 & 0.304 \\ \hline
$Vilar\:et\:al.$~\cite{10.1371/journal.pone.0058321} & 0.787 & 0.38 \\ \hline
$Average\:common\:neighbors$ & 0.802 & 0.385 \\ \hline
$Average\:Jaccard\:coefficient$ & 0.804 & 0.370 \\ \hline
$Adamic/Adar$ & 0.791 & 0.388 \\ \hline
${Katz}_b$ & 0.798 & 0.395 \\ \hline
\end{tabular}
\begin{flushleft}
\end{flushleft}
\label{table2}
\end{adjustwidth}
\end{table}

\subsection*{Comparison to multi-source data-based predictors}
In this subsection, we report the results of AMF, AMFP, the XGBoost classifier, and the methods presented by Zhang et al.~\cite{Zhang2017}, adopting the cross-validation evaluation used by Zhang et al. Table~\ref{table3} and~\ref{table4} present the results for three and five-fold cross-validation. The optimal value for AMFP's $\alpha$ is zero, hence its performance is equivalent to that of AMF. Therefore, we do not present its results in the tables. As can be seen, AMF outperforms all of the methods, including the ensembles proposed by Zhang et al. and the XGBoost ensemble. Furthermore, when adding AMF to the ensembles proposed by Zhang et al. the performance of the ensembles is still lower than that of AMF on its own. The difference between AMF and the other methods presented in table~\ref{table3} and~\ref{table4} are statistically significant. The differences are also statistically significant when comparing the XGBoost ensemble to the other methods presented in the tables. We report a $p-value<10^{-4}$ for all tests. These results which indicate that methods based on interaction networks (known DDIs) perform better than methods based on other data types align with the results presented by Zhang et al. Unfortunately, we are unable to compare the methods using retrospective analysis due to data unavailability. Cross-validation is very similar to hold-out analysis; in both cases, interactions are selected randomly and used as a test set. The differences between our hold-out analysis and our retrospective analysis indicate that if the multi-source data-based predictors and the methods we propose were compared using retrospective analysis, the differences would be even greater.

% Place tables after the first paragraph in which they are cited.
\begin{table}[H]
\begin{adjustwidth}{-2.25in}{0in} % Comment out/remove adjustwidth environment if table fits in text column.
\centering
\caption{
{\bf Area under the ROC and precision-recall curves for multi-source data comparison, three-fold cross-validation.}}
\begin{tabular}{|l|l|l|l|}
\hline
\multicolumn{1}{|l|}{\bf Method} &
\multicolumn{1}{|l|}{\bf Similarity} & \multicolumn{1}{|l|}{\bf AUROC curve} & \multicolumn{1}{|l|}{\bf AUPR curve}\\ \thickhline
AMF &  &\bf 0.9561 &\bf 0.846\\ \hline
XGB &  & 0.9544 & 0.8239\\ \hline
\multirow{14}{*}{Neighbor recommender} & Substructure & 0.9355 & 0.8079\\ \cline{2-4}
 & Target & 0.8068 & 0.4245\\ \cline{2-4}
 & Transporter & 0.7135 & 0.405\\ \cline{2-4}
 & Enzyme & 0.753 & 0.4367\\ \cline{2-4}
 & Pathway & 0.8102 & 0.6242\\ \cline{2-4}
 & Indication & 0.9034 & 0.64\\ \cline{2-4}
 & Label & 0.935 & 0.8034\\ \cline{2-4}
 & Off label & 0.9389 & 0.8153\\ \cline{2-4}
 & CN & 0.9403 & 0.8161\\ \cline{2-4}
 & AA & 0.9407 & 0.8165\\ \cline{2-4}
 & RA & 0.9423 & 0.8187\\ \cline{2-4}
 & Katz & 0.9327 & 0.7815\\ \cline{2-4}
 & ACT & 0.9099 & 0.7953\\ \cline{2-4}
 & RWR & 0.9395 & 0.8139\\ \hline
\multirow{14}{*}{Random walk} & Substructure & 0.9349 & 0.8068\\ \cline{2-4}
 & Target & 0.8442 & 0.6083\\ \cline{2-4}
 & Transporter & 0.7124 & 0.4364\\ \cline{2-4}
 & Enzyme & 0.7603 & 0.533\\ \cline{2-4}
 & Pathway & 0.8102 & 0.6477\\ \cline{2-4}
 & Indication & 0.9396 & 0.821\\ \cline{2-4}
 & Label & 0.9357 & 0.8091\\ \cline{2-4}
 & Off label & 0.9367 & 0.8116\\ \cline{2-4}
 & CN & 0.9371 & 0.8071\\ \cline{2-4}
 & AA & 0.9369 & 0.806\\ \cline{2-4}
 & RA & 0.9356 & 0.7992\\ \cline{2-4}
 & Katz & 0.9363 & 0.8012\\ \cline{2-4}
 & ACT & 0.9077 & 0.7681\\ \cline{2-4}
 & RWR & 0.9383 & 0.8128\\ \hline
Matrix perturbation method &  & 0.941 & 0.8133\\ \hline
Weighted average ensemble &  & 0.9469 & 0.8329\\ \hline
Ensemble classifier (L1) &  & 0.9537 & 0.8408\\ \hline
Ensemble classifier (L2) &  & 0.952 & 0.8391\\ \hline

\end{tabular}
\begin{flushleft}
\end{flushleft}
\label{table3}
\end{adjustwidth}
\end{table}

% Place tables after the first paragraph in which they are cited.
\begin{table}[H]
\begin{adjustwidth}{-2.25in}{0in} % Comment out/remove adjustwidth environment if table fits in text column.
\centering
\caption{
{\bf Area under the ROC and precision-recall curves for multi-source data comparison, five-fold cross-validation.}}
\begin{tabular}{|l|l|l|l|}
\hline
\multicolumn{1}{|l|}{\bf Method} &
\multicolumn{1}{|l|}{\bf Similarity} & \multicolumn{1}{|l|}{\bf AUROC curve} & \multicolumn{1}{|l|}{\bf AUPR curve}\\ \thickhline
AMF &  &\bf 0.9591 &\bf 0.8108\\ \hline
XGB &  & 0.9588 & 0.8017\\ \hline
\multirow{14}{*}{Neighbor recommender} & Substructure & 0.9362 & 0.7593\\ \cline{2-4}
 & Target & 0.8197 & 0.3642\\ \cline{2-4}
 & Transporter & 0.7143 & 0.3288\\ \cline{2-4}
 & Enzyme & 0.7562 & 0.3774\\ \cline{2-4}
 & Pathway & 0.812 & 0.5714\\ \cline{2-4}
 & Indication & 0.9119 & 0.5992\\ \cline{2-4}
 & Label & 0.9359 & 0.7537\\ \cline{2-4}
 & Off label & 0.9397 & 0.768\\ \cline{2-4}
 & CN & 0.9411 & 0.7671\\ \cline{2-4}
 & AA & 0.9414 & 0.7676\\ \cline{2-4}
 & RA & 0.9432 & 0.7704\\ \cline{2-4}
 & Katz & 0.9373 & 0.7352\\ \cline{2-4}
 & ACT & 0.9044 & 0.7239\\ \cline{2-4}
 & RWR & 0.9409 & 0.7666\\ \hline
\multirow{14}{*}{Random walk} & Substructure & 0.9356 & 0.7578\\ \cline{2-4}
 & Target & 0.8518 & 0.5599\\ \cline{2-4}
 & Transporter & 0.7127 & 0.3627\\ \cline{2-4}
 & Enzyme & 0.7609 & 0.4701\\ \cline{2-4}
 & Pathway & 0.8108 & 0.5943\\ \cline{2-4}
 & Indication & 0.9409 & 0.7771\\ \cline{2-4}
 & Label & 0.9364 & 0.7606\\ \cline{2-4}
 & Off label & 0.9374 & 0.7636\\ \cline{2-4}
 & CN & 0.938 & 0.7568\\ \cline{2-4}
 & AA & 0.9379 & 0.7556\\ \cline{2-4}
 & RA & 0.9367 & 0.7481\\ \cline{2-4}
 & Katz & 0.9374 & 0.7504\\ \cline{2-4}
 & ACT & 0.9007 & 0.7085\\ \cline{2-4}
 & RWR & 0.9392 & 0.7644\\ \hline
Matrix perturbation method &  & 0.9484 & 0.7818\\ \hline
Weighted average ensemble &  & 0.9507 & 0.7955\\ \hline
Ensemble classifier (L1) &  & 0.9571 & 0.8073\\ \hline
Ensemble classifier (L2) &  & 0.9562 & 0.806\\ \hline

\end{tabular}
\begin{flushleft}
\end{flushleft}
\label{table4}
\end{adjustwidth}
\end{table}

\section*{Discussion}
Drug interactions are the cause of many patient visits to physicians and emergency units. Estimates of the number of patients harmed due to drug interactions range from 3-5\% of all medication errors within hospitals~\cite{pmid10192758, pmid17047216}. Potential DDIs are often not discovered until the third phase of a clinical trial or in many cases, only after the drug has already been on the market for some time. In silico drug-drug interaction prediction methods, such as the methods proposed in the current research, are the most practical way of detecting DDIs.
We introduced AMF and AMFP, two new methods for in silico drug-drug interaction prediction and used DrugBank to demonstrate the superiority of the proposed methods compared to existing methods for the following metrics: AUROC and AUPR curves, precision @ n, and average precision @ n per drug. The improvement was demonstrated by predicting the interactions for a new version of DrugBank and when using a holdout evaluation scheme. We demonstrate that our methods are capable of handling both pharmacokinetic and pharmacodynamic DDIs. In addition, our results indicate that the interaction network (known DDIs) is the most useful data source for identifying potential DDIs. An ensemble method trained using XGBoost obtained better results than AMF and AMFP in most metrics and evaluation schemes. Potentially, the XGBoost ensemble can be further improved by adding more models or including domain specific information (e.g., structural data), however this may come at the cost of much longer training time. Additionally, as the multi-source data-based predictors comparison demonstrates, more data sources do not necessarily improve the performance.

\begin{figure}[H]
\includegraphics[]{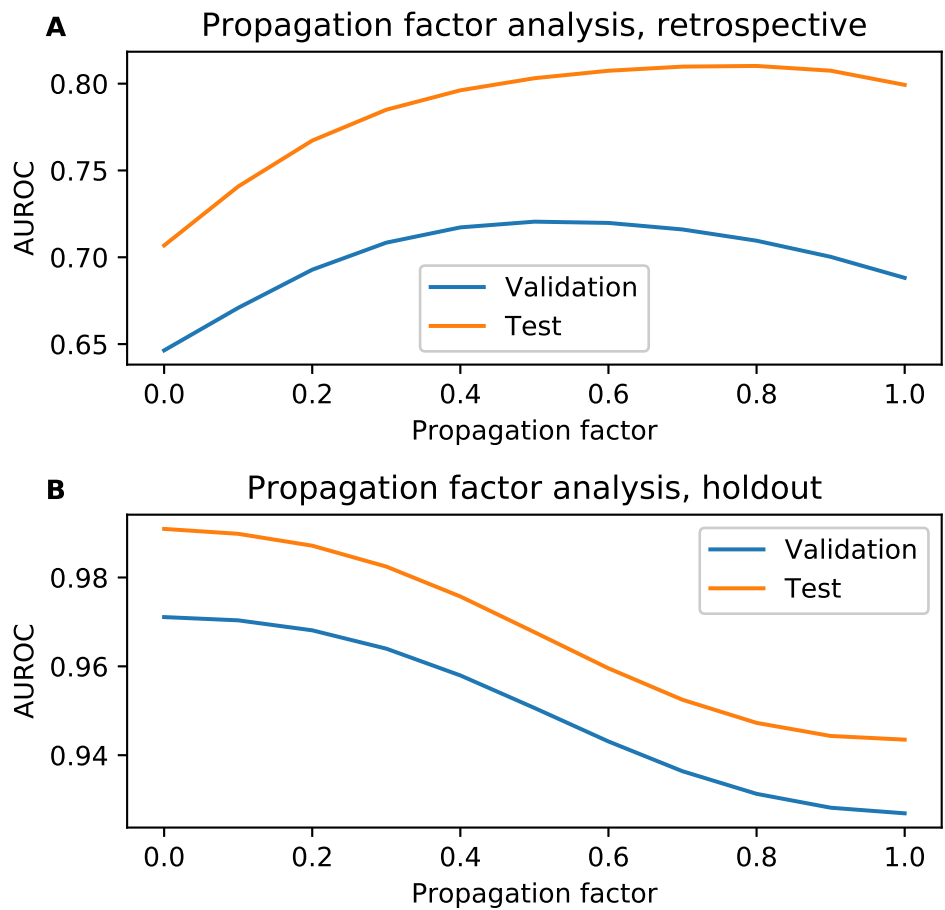} 
\caption{{\bf AMFP's propagation factor analysis.}
\textbf{A)} Retrospective propagation factor analysis. The optimal value selected during validation and used for model training is 0.5. The optimal value for the test set is 0.8. \textbf{B)} Holdout propagation factor analysis. For both validation and training the optimal value is zero - weights are not propagated at all. } 
\label{fig5}
\end{figure}

In this section, we present a more in depth analysis on the propagation factor of AMFP, testing different values for the validation and test sets to investigate whether there is a strong correlation between the two. Figure~\ref{fig5} presents AMFP's propagation factor analysis for the hold-out and retrospective analysis. A big difference can be seen between the two evaluation schemes. For the retrospective scheme the optimal values are 0.5 and 0.8 respectively for the validation and test sets. The effect of different propagation factor values on the AUROC of the validation and test sets is similar. For the holdout evaluation scheme the effect of different propagation factor values on the AUROC of the validation and test sets is very similar; smaller values are preferred, and zero is the optimal value. This means that no propagation is required for the validation or test sets. This difference could be the result of the difference in the test set distributions. As stated in the \nameref{Evaluation} section, retrospective analysis is preferred as it is more true to life and necessitates that the model generalizes better than the holdout analysis. Both evaluation techniques show better results on the test set, which is relatively unusual. For the holdout analysis the difference is very small (about 0.02); this difference might simply be explained by the amount of data, in that the model evaluated on the validation set is trained using less data than the model evaluated on the test set. For the retrospective analysis where the difference is larger, the reason is probably similar: the previous version of DrugBank used for training during validation contains fewer interactions (11,284 interactions) than the version used to train the final model which was used for testing (with 45,296 interactions). 
We report the optimal values for the parameters used: the embedding size used are 256 and 512 for the retrospective and holdout analysis respectively. For both evaluation schemes, the dropout is 0.3, and the learning rate is 0.01. It's important to note that the dropout is applied separately on the embedding layer of each of the drugs. On average the number of entries in the embedding vector which are not affected at all by dropout during training is given by:
\begin{equation}
k\cdot(1-p)^2,
\end{equation}
where $k$ is the embedding size, and $p$ is the dropout ratio. Hence, for the retrospective analysis where the embedding size was 512 on average only 184.32 embedding entries are unaffected by the dropout during training. The optimal number of epochs used is five and six for the retrospective and holdout analysis respectively, and the batch sizes are 1024 and 256 respectively. For the comparison to multi-source data-based predictors the optimal values for the parameters used in AMF were: embedding size of 64, dropout of 0.5, forty epochs, batch size of 256 and a learning rate of 0.01.
Some interesting observations can be made by comparing AMFP's predictions regarding each drug pair with the pair's structural similarity. One can expect that if two drugs share similar interactions it is likely that they have some structural similarity, however the predictions and structural similarity might be different or complementary. We computed the correlation coefficient between AMFP's predictions and the structural similarity. For the structural similarity we used the method which was used by Ryu et al.~\cite{RyuE4304}. This comparison showed a low correlation coefficient of 0.151 with $P-value<10^{-15}$; for comparison, Vilar et al.~\cite{10.1371/journal.pone.0058321} report a correlation coefficient of 0.167.

\subsection*{Practical contribution}
The proposed models can be utilized to improve drug-drug interaction discovery and can be combined with additional structural information to improve drug-drug interaction detection performance. 

In this paper, we present the top 100 predictions made by AMFP (see \nameref{Supporting information} section) after training it on release 5.1.1 of DrugBank and using the parameters optimized in the retrospective analysis; note that version 5.1.1 was the latest release available when our research was conducted. We manually validated the first 10 predictions made by AMFP; as of now, eight of them have been added to DrugBank. For the following five drug-drug interactions the metabolism of one drug can change due to the drug interaction: Curcumin and Primidone, Rifapentine and Fluvoxamine, Curcumin and Rifapentine, Lumacaftor and Fluvoxamine, and Curcumin and Lovastatin. For the following three drug-drug interactions the serum concentration of one drug can change due to the drug interaction: Ceritinib and Fluvoxamine, Curcumin and Clotrimazole, and Curcumin and Lumacaftor. No evidence was found for the existence of the following two drug-drug interactions: Pentobarbital and Sulfisoxazole, and Curcumin and Pentobarbital which might indicate two new unknown interactions predicted by our methods.

In addition, we created a list of the latent factors (embeddings) created for each drug and made this publicly available. These factors can be used as compressed representations of the drugs. The factors contain the structure of the interaction network and can provide a head start on downstream tasks in the form of transfer learning. For example, the drug embeddings created using the interaction network can be used to detect side effects.

AMF and AMFP can be scaled to support a large number of drugs and interactions; the models do not require training using all of the positive examples (existing interactions), and positive sampling can also be used, allowing the method to operate on very large datasets. Each of the proposed methods presented here required no more than a few minutes to train using a standard laptop. Hyperparameter optimization required more time, and this process can usually be executed efficiently by an expert; AutoML methods for parameter optimization are currently gaining interest, and such methods can dramatically reduce the time required for optimizing hyperparameters~\cite{8477921}.

\section*{Conclusion}
In this paper, we designed two methods for drug-drug interaction prediction based on a novel matrix factorization technique designed for adjacency matrices and developed useful in silico models to predict new drug interactions. Additionally, we train an XGBoost ensemble using various predictors. The methods were implemented and made public, along with additional resources used in this research. 
Our methods were systematically validated through a retrospective and holdout evaluation using DrugBank (release 5.1.1 which contains 1,440 drugs and 248,146 drug-drug interactions), showing state of the art results with an area under receiver operating characteristic curve of 0.814 overall and accuracy of 56\% when predicting the first interaction for each drug. Additionally, we compare and demonstrate the superiority of our methods over existing state-of-the-art methods, which were trained using various data sources, using cross-validation. Our methods can be used on a large-scale and applied for link prediction problems in domains other than drug-drug interaction prediction. 
Using the proposed DDI predictor, a database containing the most promising drug-drug interaction candidates is provided in the \nameref{Supporting information} section.

\section*{Supporting information} \label{Supporting information}
\paragraph*{S1 Appendix.}
\label{S1_Appendix_AMF}
{\bf AMFP’s predictions: a list of 100 non-existing drug-drug interactions predicted by AMFP using a retrospective model on release 5.1.1 of DrugBank. The items are listed in descending order, based on the level of confidence.} Curcumin and Primidone, Ceritinib and Fluvoxamine, Rifapentine and Fluvoxamine, Curcumin and Pentobarbital, Curcumin and Rifapentine, Lumacaftor and Fluvoxamine, Curcumin and Lovastatin, Pentobarbital and Sulfisoxazole, Curcumin and Clotrimazole, Curcumin and Lumacaftor, Primidone and Atomoxetine, Phenobarbital and Atomoxetine, Idelalisib and Fluvoxamine, Fluconazole and Fluvoxamine, Procaine and Methadone, Deferasirox and Primidone, Ceritinib and Clotrimazole, Stiripentol and Rifapentine, Stiripentol and Sulfisoxazole, Curcumin and Cobicistat, Cimetidine and Ziprasidone, Desipramine and Cyclosporine, Lumacaftor and Rifapentine, Curcumin and Pimecrolimus, Cobicistat and Pentobarbital, Nefazodone and Leflunomide, Curcumin and Fluconazole, Osimertinib and Sulfisoxazole, Pentobarbital and Atomoxetine, Ceritinib and Clemastine, Doxycycline and Fluvoxamine, Clemastine and Sulfisoxazole, Lumacaftor and Clotrimazole, Lumacaftor and Stiripentol, Stiripentol and Cobicistat, Curcumin and Clemastine, St. John's Wort and Primidone, Quinidine and Tranylcypromine, Clemastine and Clotrimazole, St. John's Wort and Fluvoxamine, Cobicistat and Sulfisoxazole, Deferasirox and Pentobarbital, Nefazodone and Torasemide, Sarilumab and Fluvoxamine, Idelalisib and Sulfisoxazole, Teriflunomide and Nefazodone, Selegiline and Quinidine, Acetyl sulfisoxazole and Pentobarbital, St. John's Wort and Fosphenytoin, Selegiline and Azelastine, Ketoconazole and Tranylcypromine, St. John's Wort and Phenobarbital, Curcumin and St. John's Wort, Cobicistat and Clotrimazole, Rifapentine and Thioridazine, Desipramine and Isradipine, Nefazodone and Irbesartan, Osimertinib and Clotrimazole, Sarilumab and Primidone, Dasatinib and Desipramine, Lopinavir and Tranylcypromine, Rifapentine and Clemastine, Idelalisib and Clotrimazole, Clarithromycin and Metoprolol, Sarilumab and Phenobarbital, Ceritinib and Deferasirox, Deferasirox and Sulfisoxazole, Diphenhydramine and Verapamil, Luliconazole and Fluvoxamine, Tranylcypromine and Indinavir, Deferasirox and Rifapentine, Sarilumab and Fosphenytoin, Rotigotine and Escitalopram, Metoprolol and Sulfisoxazole, Lumacaftor and Cobicistat, Nilotinib and Tranylcypromine, Isoniazid and Ziprasidone, Thioridazine and Rifabutin, Lumacaftor and Clemastine, Simeprevir and Sulfisoxazole, Clotrimazole and Fluconazole, Curcumin and Doxycycline, Venlafaxine and Torasemide, St. John's Wort and Pentobarbital, Nefazodone and Valsartan, Teriflunomide and Phenobarbital, Teriflunomide and Venlafaxine, Stiripentol and Fluconazole, Osimertinib and Clemastine, Darunavir and Tranylcypromine, Desipramine and Conivaptan, Itraconazole and Desipramine, Cobicistat and Clemastine, Desipramine and Rifabutin, Nefazodone and Ciprofloxacin, Rifampicin and Paroxetine, Phenytoin and Betaxolol, Vemurafenib and Nabilone, St. John's Wort and Sulfisoxazole, Rifapentine and Atomoxetine.

\nolinenumbers

\bibliography{DDIP}{}
\end{document}